\documentclass[]{sjtu_sai}
\usepackage[toc,page,header]{appendix}
\usepackage{minitoc}
\usepackage{algorithm}
\usepackage{algpseudocode}
\usepackage{booktabs}
\usepackage{multirow}
\usepackage{siunitx}
\usepackage{makecell}
\usepackage{mathtools}
\usepackage{tabularx,array}
\usepackage[table]{xcolor}
\usepackage{graphicx}
\usepackage{wrapfig}
\usepackage{xspace}
\usepackage{amsmath}
\usepackage{amsfonts}
\usepackage{fontawesome5}
\usepackage{wrapfig}
\usepackage{graphicx} 
\renewcommand{\footnoterule}{%
  \kern-3pt
  \hbox{\color{saibg}\rule{0.45\columnwidth}{0.6pt}}%
  \kern2.4pt
}
\makeatletter
\renewcommand{\@makefntext}[1]{%
  \parindent0pt\noindent\@makefnmark\hspace{0.4em}#1%
}
\makeatother

\title{Grounded Semantic Re-Binding for Robust Instruction Generalization in Vision-Language-Action Models}

\author[1,2]{Zhaokai Yin}
\author[1,\dagger]{Zhipeng Zhang}
\affiliation[1]{AutoLab, School of Artificial Intelligence, Shanghai Jiao Tong University}
\affiliation[2]{Research Lab, Anyverse Dynamics}
\contribution[\dagger]{Corresponding authors}

\abstract{Vision-Language-Action (VLA) models excel in robotic manipulation but suffer catastrophic performance drops when canonical instructions are simply paraphrased. Although this brittleness is typically addressed through costly data scaling, our probing reveals that the root cause is architectural rather than a lack of semantic understanding. Specifically, we demonstrate that current VLAs successfully retain the correct task identity internally. The failure actually stems from the joint encoding of dynamic visual observations and text, which introduces systematic feature shifts. Because the downstream action policy is highly vulnerable to these variations, it fails to translate the preserved semantics into correct control commands. To resolve this structural bottleneck, we propose Grounded Semantic Re-binding (GSR), an elegant intervention that bypasses unstable joint routing by explicitly fusing independently extracted task semantics with native visual features to train a completely re-initialized action expert from scratch. This targeted intervention dramatically restores paraphrastic invariance using only canonical demonstrations. On the LIBERO-Para benchmark, GSR improves success rates by up to 44.6 percent. It enables lightweight models to rival massively scaled baselines and pushes state-of-the-art models to a new record PRIDE score of 70.4, outperforming the recently introduced large-scale pretrained model Xiaomi-Robotics-0 in instruction generation capabilities. Building on these insights, we also introduce ParaVLA, a natively decoupled 0.33B-parameter model exhibiting near-perfect robustness to instruction rewording. Ultimately, our work proves that robust semantic grounding can be achieved through elegant structural design, bypassing the inefficient brute-force data scaling paradigm.}

\checkdata[\textcolor{saibg}{\raisebox{0pt}[0pt][0pt]{\scalebox{1.15}{\faGithub}}}]{\url{https://github.com/AutoLab-SAI-SJTU/GSR-ParaVLA}}
\begin{document}
\maketitle%
\begingroup
\renewcommand{\thefootnote}{*}
\footnotetext{Work was done during the internship at AutoLab, SAI, SJTU.}
\endgroup

\section{Introduction}
\label{sec:intro}
\begin{wrapfigure}{r}{0.48\textwidth}
    \vspace{-0.8\baselineskip}
    \centering
  \includegraphics[width=1 \linewidth]{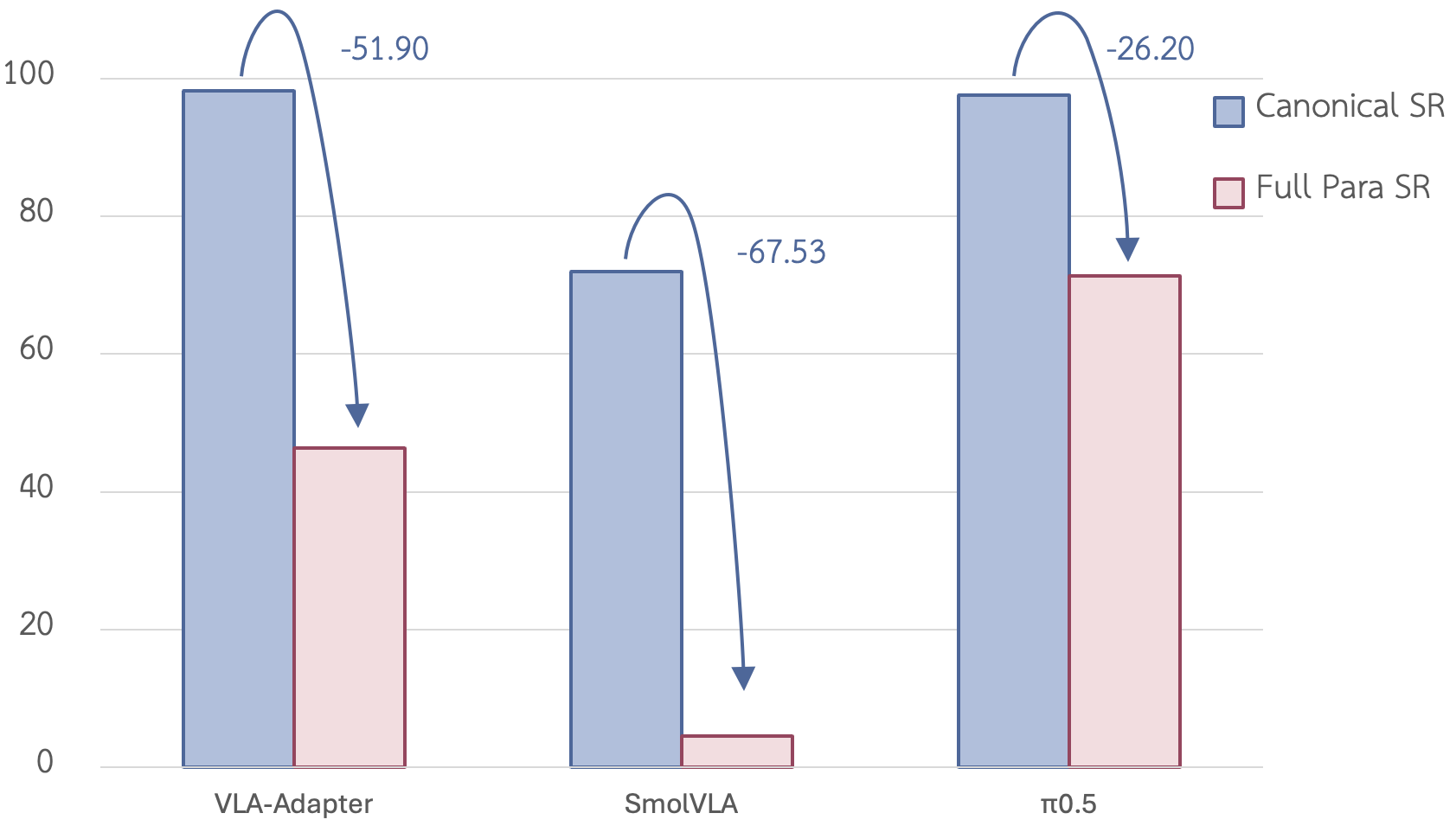}
    \caption{Equivalent instruction rewrites cause substantial success drops although the physical tasks remain unchanged. }
    \label{fig:Fig0}
    %\vspace{-1em}
\end{wrapfigure}

Vision-language-action (VLA) models have demonstrated impressive capabilities across diverse manipulation tasks by mapping visual perception and language instructions to robot actions~\cite{rt1,rt2,openvla,palme,pi0,pi05,gr00tn1,vla-adapter,xvla,openxe}. However, this apparent success heavily relies on the rigid, canonical language templates predominantly used in standard benchmarks. When faced with equivalent paraphrased instructions, current VLAs exhibit a critical vulnerability and fail to follow the underlying semantic meaning. As illustrated in Fig.~\ref{fig:Fig0}, state-of-the-art models such as VLA-Adapter~\cite{vla-adapter}, SmolVLA~\cite{smolvla}, and $\pi_{0.5}$~\cite{pi05} suffer catastrophic performance drops of up to 67.53\% when canonical instructions are simply reworded. 

\enlargethispage{\baselineskip}
To investigate this discrepancy, we evaluate VLAs through the lens of paraphrastic invariance, where equivalent task descriptions must yield consistent robot behavior~\cite{applesofa,groundingvla,stablelanguageguidance,liberopara,bcz,cliport}. Surprisingly, our controlled analysis reveals that models like VLA-Adapter, SmolVLA, and $\pi_{0.5}$ successfully retain the correct task identity internally even when final execution fails (see Sec.~\ref{sec:prob}).

Under identical visual observations, the internal features encoded from paraphrased instructions still predominantly align with their canonical counterparts. This paradox suggests that the breakdown occurs not in understanding the instruction, but in how the encoded information is utilized during action generation.

A common strategy to mitigate these failures in paraphrastic invariance involves massively broadening language coverage during training. While this data-driven approach improves robustness, it incurs immense computational costs. Furthermore, our previous analysis demonstrates that VLAs already encode the correct task semantics internally (detailed in Sec.~\ref{sec:prob}). This realization suggests that brute-force data augmentation is an inefficient way to fix a purely architectural translation issue. We therefore propose an orthogonal solution.

To formulate this orthogonal solution, we must first pinpoint exactly where this architectural translation fails. We perform a targeted token intervention on VLA-Adapter. During a paraphrase execution, we isolate the final action block and replace its language tokens with those extracted from a canonical instruction run. By keeping the visual and state tokens completely unchanged, this precise substitution dramatically boosts the execution success rate from 60\% to 96\%. Motivated by this insight, we introduce Grounded Semantic Re-binding (GSR), a simple yet highly effective architectural intervention that explicitly repairs this broken pathway using only canonical demonstrations. Rather than relying on the unstable native language routing of VLAs, GSR extracts robust instruction semantics using a frozen T5 encoder~\cite{t5} and directly injects them into the architecture's native visual and state representations. To bypass the original flawed utilization mechanism entirely, we re-initialize the model's action expert rather than fine-tuning its pre-trained weights. By combining stable external task semantics, native scene grounding, and a relearned action mapping, GSR successfully circumvents these architectural failure modes and restores robust instruction following.

Without any paraphrase training, GSR significantly improves success rates on the full 4,092-episode LIBERO-Para~\cite{liberopara} evaluation, which preserves the physical task while modifying action expressions and object references. Success increases dramatically from 46.82\% to 70.94\% on VLA-Adapter and from 4.47\% to 49.12\% on SmolVLA, all while preserving canonical-task ability. Remarkably, this simple architectural intervention elevates the instruction-following capability of VLA-Adapter, a model without massive pre-training, to a level comparable with heavily scaled models like $\pi_{0.5}$. For $\pi_{0.5}$ itself, the baseline performance is already exceptionally high (73.60\%) due to its large-scale heterogeneous co-training and robust language generalization. Yet, GSR still pushes this performance to 75.59\% and achieves the highest reported PRIDE score of 70.4 (\textit{vs.} recent Xiaomi-Robotics 69.2). This continued improvement on a heavily scaled model illustrates the orthogonality of our approach. It demonstrates that robust semantic grounding can be achieved efficiently, allowing lightweight models to rival massive ones without requiring exhaustive paraphrase data or costly joint pre-training recipes. Beyond retrofitting existing VLAs, we further present ParaVLA, a 0.33B-parameter model featuring a natively decoupled architecture that separates language and vision encoding. This model demonstrates the potential of such a simple structure to achieve remarkable paraphrastic invariance, exhibiting a negligible performance gap between canonical and paraphrased language instructions. 

\enlargethispage{\baselineskip}
In summary, our main contributions are threefold.

\begin{itemize}
\item We identify the root cause of paraphrastic invariance failures in VLAs. Our probing reveals that models successfully encode task semantics internally but fail due to a flawed architectural mechanism translating these representations into control commands.

\item We propose Grounded Semantic Re-binding. By fusing stable T5 semantics with native visual features and training a completely re-initialized action expert from scratch, GSR explicitly repairs the broken semantic pathway without requiring exhaustive paraphrase data.

\item We demonstrate the universality of GSR across diverse architectures. It enables lightweight models like VLA-Adapter to rival massive baselines and further improves heavily scaled models like $\pi_{0.5}$. GSR achieves the highest reported PRIDE score of 70.4, proving its structural benefits are orthogonal to massive data scaling.
\end{itemize}

%\clearpage
\section{Analysis}
\label{sec:analysis}
\subsection{Observation}
Our analysis begins with a striking observation regarding the extreme fragility of instruction following in VLAs. We train the action modules of evaluated policies on canonical LIBERO-Goal demonstrations and evaluate them using the LIBERO-Para protocol~\cite{libero}. This protocol strictly preserves all physical conditions and success criteria, altering only the textual instructions through unseen rewrites of action expressions or object references. Since the required physical robot behavior remains entirely unchanged, a reliable policy should theoretically yield similar success rates across both conditions. Instead, Fig.~\ref{fig:Fig0} reveals a catastrophic performance drop across all evaluated models. This vulnerability is particularly surprising for VLA-Adapter. Because it was explicitly designed to minimize fine-tuning of the VLM backbone, a natural intuition is that it should preserve the VLM's inherent robustness to linguistic variations. Its severe degradation in such a straightforward setting highlights that standard VLAs heavily overfit to the training syntax, proving that simply retaining a powerful VLM backbone is insufficient to guarantee paraphrastic robustness.

This phenomenon raises a fundamental question. We must determine whether the unseen rewrites completely corrupt the model's internal semantic representation, making the requested task indistinguishable from others, or whether the correct task information actually survives but is utilized inconsistently by the action generation module. The following probing analysis directly tests these two hypotheses. 

\subsection{Probing Task under Paraphrasing}
\label{sec:prob}
To locate the exact source of paraphrase failures, we first investigate whether a rewritten instruction still preserves enough semantic information for the model to distinguish the intended task. We select VLA-Adapter as our primary subject because its pretrained Qwen~\cite{qwen} backbone undergoes minimal fine-tuning, leaving most robot-specific learning to a separate Bridge-Attention policy~\cite{bridgeattention}. This architectural separation allows us to examine the model's task comprehension independently of its action generation. To strictly isolate the effect of language, we conduct our analysis on LIBERO-Goal, whose ten tasks share an identical visual environment, thereby forcing the model to rely entirely on linguistic input to identify the task. 

For the $n$-th test case, let $y_n$ denote its ground-truth task label. We define $a^p_n$ as the flattened action chunk generated from the paraphrased instruction. To determine if the model retains task-discriminative information, we evaluate the canonical instructions of all $N$ tasks under the exact same observation as case $n$. Let $a^c_{n,k}$ denote the action chunk produced by the canonical instruction of task $k$ ($1 \leq k \leq N$). To measure the behavioral similarity between these outputs, we first normalize the action vectors using a mean action $\mu$ estimated from a canonical reference corpus. The normalized action $\bar{a}$ and the distance metric $D(a, b)$ are defined as

\begin{equation}
\bar{a} = \frac{a - \mu}{\max(\|a - \mu\|_2, 10^{-8})}.
\end{equation}
\begin{equation}
D(a, b) = \|\bar{a} - \bar{b}\|_2. 
\end{equation}
Using this distance, we compute the Retrieval@1 score to determine if the action produced by a paraphrase is closest to the action produced by the correct canonical instruction among all $N$ candidates. The metric is formulated as
\begin{equation}
\text{Retrieval@1} = \frac{1}{M} \sum_{n=1}^{M} \mathbb{I} \left[ \arg\min_{1 \leq j \leq N} D(a^p_n, a^c_{n,j}) = y_n \right]. 
\end{equation}

Furthermore, we evaluate a wrong-task instruction for each case to establish a baseline distance. Let $\Delta_{\text{para}}$ and $\Delta_{\text{wrong}}$ represent the average distances from the canonical output to the paraphrase output and the wrong-task output, respectively. We quantify semantic retention $R$ as
\begin{equation}
R = \frac{\Delta_{\text{wrong}} - \Delta_{\text{para}}}{\Delta_{\text{wrong}} + \Delta_{\text{para}}}. 
\end{equation}

\begin{wraptable}{r}{0.48\textwidth}
\vspace{0pt}
\centering
\small
\setlength{\tabcolsep}{3.5pt}
\renewcommand{\arraystretch}{1.08}
\begin{tabular}{@{}lccc@{}}
\toprule
Model & Retrieval@1$\uparrow$ & $R$$\uparrow$ &
$\Delta_{\mathrm{w}}/\Delta_{\mathrm{p}}$ \\
\midrule
VLA-Adapter & 0.675 & 0.467 & 2.752 \\
SmolVLA     & 0.516 & 0.303 & 1.872 \\
$\pi_{0.5}$ & 0.941 & 0.604 & 4.051 \\
\bottomrule
\end{tabular}
\caption{Task information in predicted actions under paraphrases. The last column reports the wrong-task to paraphrase distance ratio.}
\label{tab:task_retention1}
\vspace{-0.5\baselineskip}
\end{wraptable}

A positive $R$ indicates that the model's response to a paraphrase remains closer to the intended task than to an entirely different task. Tab.~\ref{tab:task_retention1} summarizes this analysis for all three models. All three obtain above-chance Retrieval@1 scores, positive $R$ values, and larger wrong-task than paraphrase distances. While these aggregate metrics do not imply perfect comprehension in every failed episode, they demonstrate that the performance drop cannot be explained by a complete loss of task semantics.

To causally verify whether this retained internal information is sufficient to correct the robot's behavior, we perform a targeted intervention. We run both a canonical instruction and its paraphrase under identical environmental conditions. During the paraphrase execution, we intervene right before the final Bridge-Attention block. We replace only the Qwen output feature entering this block with its canonical counterpart, leaving all other inputs untouched. Remarkably, this single feature replacement eliminates 96.8\% of the discrepancy in the predicted actions and raises the paired success rate from 60\% to 96\%. This intervention causally proves that the correct task information survives the language backbone but is severely mishandled by the downstream action policy, directly motivating our architectural redesign.

\subsection{Instability of Joint V-L Encoding}

The previous causal intervention confirms that replacing the Qwen feature with its exact canonical counterpart corrects the robot's behavior.

\begin{wrapfigure}{r}{0.46\textwidth}
\vspace{0pt}
\centering
\includegraphics[width=\linewidth]{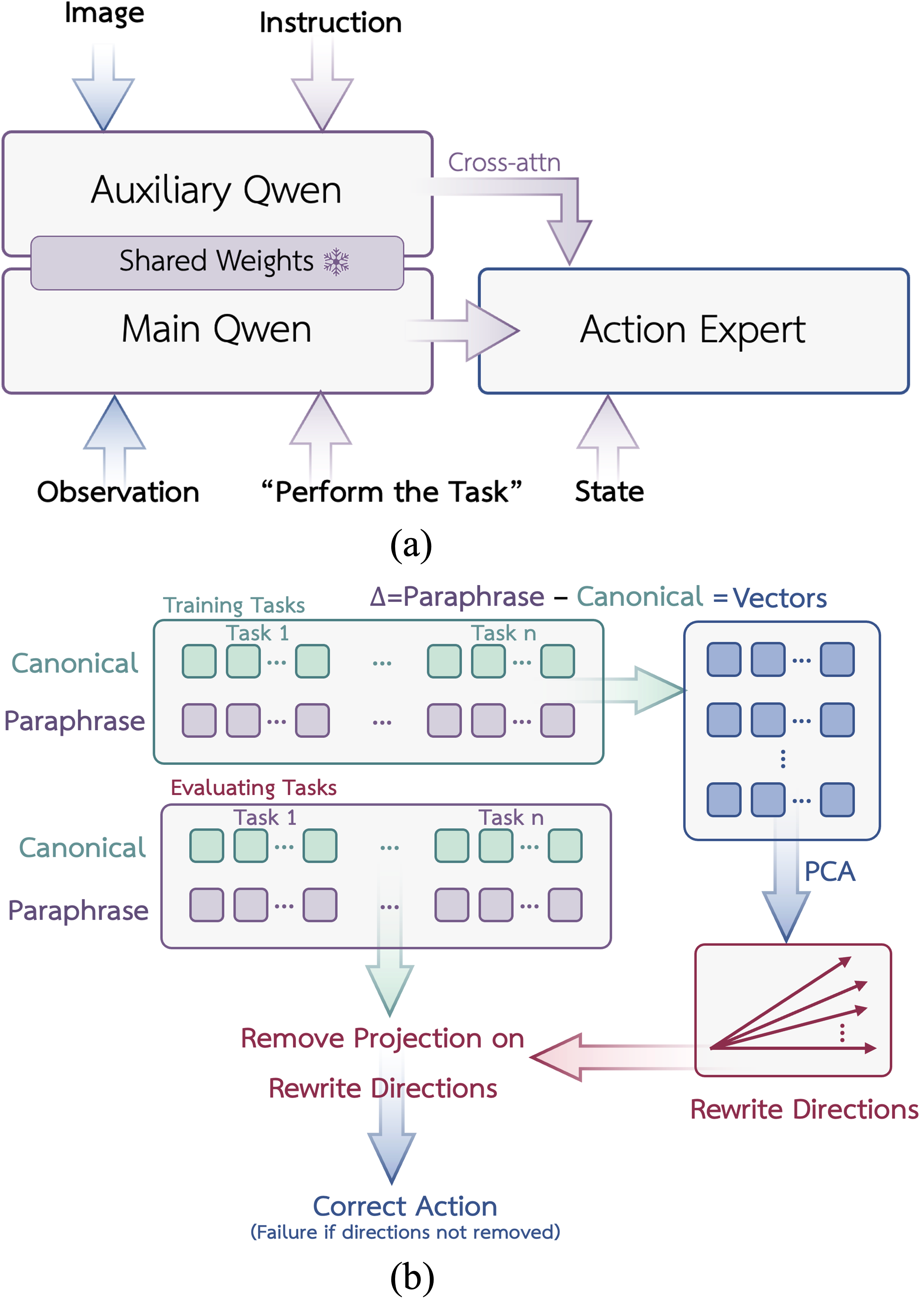}
\caption{Visual and wording controls: \textbf{(a)} auxiliary-image replacement; \textbf{(b)} task-disjoint rewrite-direction removal.}
\label{fig:Fig01}
\vspace{-1\baselineskip}
\end{wrapfigure}

This reveals a crucial nuance that while our probing analysis showed that paraphrased features preserve the correct task identity, they are not strictly identical to the canonical features. Equivalent instructions still produce different feature outputs at this layer, and these differences are what ultimately alter the final action. This raises the question: \textit{why} do semantically equivalent instructions produce different outputs here? In the VLA-Adapter architecture, the auxiliary Qwen branch processes the instruction jointly with the current visual observation. We test whether either the dynamic image or the wording itself introduces the variations that change the action.

\textbf{Influence of Visual on Task Representations:} First, as shown in Fig.~\ref{fig:Fig01}(a), we isolate the effect of the visual input. We replace only the image received by the auxiliary Qwen branch with a fixed dummy image across all episodes, while the main vision pathway retains the real, dynamic observation. Strikingly, this simple decoupling raises the full Para success rate from 46.82\% to 61.58\% (a 14.76-point paired gain). We also verify that using a fixed natural image yields a similar 7.17-point improvement in paired paraphrase success, showing that the result is not specific to one replacement image. This indicates that jointly encoding the dynamic visual input exacerbates the feature variations caused by paraphrasing.

\textbf{Influence of Wording on Task Representations:} Next, as shown in Fig.~\ref{fig:Fig01}(b), we isolate the variation caused strictly by wording. Using a 5-fold task-disjoint cross-validation, we use eight tasks to estimate 32 specific feature directions that consistently separate the Qwen outputs of canonical instructions from their paraphrases. We then remove the components along these directions from the Qwen outputs of the remaining two unseen tasks. Here, we measure the action gap as the mean relative root-mean-square difference between the flattened action chunks predicted for the canonical and paraphrased instructions. Removing these 32 estimated directions significantly reduces this action gap from 0.4361 to 0.2282, whereas removing 32 random directions (with the same norm) leaves the gap at 0.4386. Furthermore, on 20 independent closed-loop episodes, removing these estimated directions raises the success rate from 55\% to 90\%. This result clearly demonstrates that wording changes do not destroy the underlying task semantics, but rather introduce a systematic, separable shift in the feature space. The downstream action policy is highly vulnerable to this specific linguistic variation, failing to extract the invariant task identity. Note that this intervention is used purely to diagnose the effect of wording variation, as estimating these directions requires paired canonical and paraphrased instructions.

Importantly, because the real observation remains available through the main visual pathway, these control experiments do not remove visual grounding. Instead, they demonstrate that jointly encoding the current image and the task wording in the auxiliary input is the root cause of the feature variations that degrade the action. This insight directly motivates our proposed strategy: to extract the task meaning independently of the current image, and only relate it to the visual scene later inside the action policy.

\section{Solution and Experiments}
\label{sec:solution}

\subsection{Repairing VLA-Adapter}
Based on the preceding analysis, we require an instruction-only semantic source that is more stable under equivalent wording than the native task input. We instantiate this source with frozen T5-large. Qwen continues to process the real auxiliary image, but receives the same fixed sentence, “perform the task,” for every training and evaluation sample. It therefore supplies no task-specific wording. A learned projection maps the T5 output to the condition width used by the Bridge-Attention policy. The policy then combines this instruction-derived condition with the real visual input and robot state at every block.

Like standard VLA-Adapter training, the Bridge-Attention policy is trained from scratch. This allows the policy to learn from the beginning of training how the T5 condition, visual features, and robot state jointly determine the action, rather than retaining an action mapping fitted to the original language input.  T5 and the pretrained multimodal backbone remain frozen, while the projection and the action policy are optimized with the original action loss. All GSR variants are trained only on the ten canonically instructed tasks in LIBERO-Goal. No LIBERO-Para rewrite or manually generated paraphrase is used during training, and each trained configuration uses a fixed training seed.

We first evaluate on LIBERO-Para, which preserves tasks, initial states, and success criteria while rewriting action expressions, object references, or both. Full Para comprises 4,092 episodes (870 Act, 259 Obj, 2,963 Comp). Each paired comparison shares the episode, initial state, success criterion, and action-generation seed.

% We first evaluate the model on LIBERO-Para, which preserves the task, initial state, and success criterion while rewriting the action expression, object reference, or both. Full Para contains 4,092 episodes, including 870 Act, 259 Obj, and 2,963 Comp episodes. Each paired comparison uses the same episode, initial state, success criterion, and action-generation seed. 

Tab.~\ref{tab:main_results} shows that GSR substantially improves the language generalization of VLA-Adapter ($46.82\% \rightarrow 70.94\%$ ), whereas simply adding T5 while retaining the original Qwen instruction provides little improvement ($46.82\% \rightarrow 47.31\%$). The gain therefore comes from changing how task information is provided to action generation, rather than from merely adding another language encoder.

%%%%%%%%%%%%%%%%%%%%%%%%%%%%大表Begin
\begin{table*}[t]
\centering
\small
\setlength{\tabcolsep}{5.2pt}
\renewcommand{\arraystretch}{1.12}
\begin{tabular*}{\textwidth}{
    @{\extracolsep{\fill}}
    l l c c c c
    @{}
}
\toprule
Model
& Configuration
& Goal SR $\uparrow$
& Full Para SR $\uparrow$
& Drop $\downarrow$
& PRIDE $\uparrow$ \\
\midrule

OpenVLA-OFT$_{\mathrm{Goal}}^\dagger$~\cite{openvla-oft}
& Reported
& 97.9 & 64.7 & 33.2 & 58.8 \\

OpenVLA-OFT$_{\mathrm{Mixed}}^\dagger$
& Reported 
& 96.1 & 63.7 & 32.4 & 56.3 \\

\midrule

SmolVLA~\cite{smolvla}
& Native
& 72.0 & 4.47 & 67.53 & 2.6 \\

SmolVLA
& Native + T5
& 76.0 & 13.49 & 62.51 & 7.9 \\

\textbf{SmolVLA}
& \textbf{GSR}
& \textbf{78.0}
& \textbf{49.12}
& \textbf{28.88}
& \textbf{41.4} \\

\midrule

X-VLA$^\dagger$~\cite{xvla}
& Reported
& 97.8 & 62.1 & 35.7 & 52.7 \\

\midrule

VLA-Adapter$^\dagger$~\cite{vla-adapter}
& Reported
& 98.2 & 46.3 & 51.9 & 36.1 \\

VLA-Adapter
& Native 
& 98.2 & 46.82 & 51.38 & 36.7 \\

VLA-Adapter
& Native + T5
& 97.0 & 47.31 & 49.69 & 37.1 \\

\textbf{VLA-Adapter}
& \textbf{GSR}
& \textbf{98.0}
& \textbf{70.94}
& \textbf{27.06}
& \textbf{62.0} \\

\midrule

Xiaomi-Robotics-0$^\dagger$~\cite{xiaomi}
& Reported
& 98.8 & \textbf{76.0} & 22.8 & 69.2 \\

\midrule

$\pi_{0.5}^\dagger$~\cite{pi05}
& Reported
& 97.6 & 71.4 & 26.2 & 65.4 \\

%$\pi_{0.5}^\dagger$
%& Expert-only
%& 78.6 & 39.1 & 39.5 & 32.0 \\

$\pi_{0.5}$
& Native
& 93.0 & 73.60 & 19.40 & -- \\

\textbf{$\pi_{0.5}$}
& \textbf{GSR}
& \textbf{91.0}
& \textbf{75.59}
& \textbf{15.41}
& \textbf{70.4} \\

\textbf{$\pi_{0.5}$}
& \textbf{GSR}$^{*}$
& \textbf{96.0}
& \textbf{75.76}
& \textbf{20.24}
& \textbf{70.3} \\

\bottomrule
\end{tabular*}
\caption{Language-generalization results on LIBERO-Para. Rows marked $\dagger$ are reported by LIBERO-Para. Native denotes the task-specific instruction given to the model's original VLM. Native + T5 retains this input while adding T5.  Drop is Goal SR minus Full Para SR. Bold rows are GSR variants described in Sec.~\ref{sec:solution}. $^{*}$ Training volume doubled.}
\label{tab:main_results}
%\vspace{-1em}
\end{table*}

% Matched rows use the same training data, action-module initialization rule, and evaluation episodes.
%%%%%%%%%%%%%%%%%%%%%%%%%%%%大表End

\subsection{Extending the Design beyond VLA-Adapter}
\label{sec:4.2}
Having improved VLA-Adapter, we next examine whether the same design extends to models with different information flow and training. SmolVLA is a compact policy in which SmolVLM~\cite{smolvlm} combines the instruction and image before a flow-matching action expert~\cite{fm}. $\pi_{0.5}$ uses a much larger PaliGemma~\cite{paligemma} backbone, heterogeneous robot data, and a flow-matching expert. Together with VLA-Adapter, these models cover a separate action policy, early vision-language interaction, and large-scale multimodal co-training.

To determine the optimal way to utilize these stable language features, we first attempt the same adaptation way as VLA-Adapter on SmolVLA. However, this approach completely fails to improve paraphrase generalization. While it preserves a 76\% success rate on canonical instructions, it achieves a mere 13.49\% on paraphrases. This catastrophic drop implies that simply forwarding isolated language features to the action head deprives the model of essential visual grounding. To address this, we alter the injection point. Instead of bypassing the visual processing, we inject the T5 output directly into SmolVLM's original language input positions. This allows the pure semantic features to deeply interact with the current image through the VLM's existing computation layers before action generation. 
This early integration drastically boosts paraphrase success to 49.12\%, while maintaining a 78\% canonical success rate. A similar design is used for $\pi_{0.5}$, where the T5 condition is introduced into the PaliGemma instruction representation before the subsequent multimodal computation. The only difference is that PaliGemma continues to receive the real task instruction instead of a fixed neutral prompt. Its native instruction input already generalizes reliably to paraphrases and remains the primary source of task control. When PaliGemma is given an instruction for a different task, success decreases from 75.59\% to 2.88\%, whereas changing only the T5 instruction leaves 74.0\% success. We therefore retain the reliable PaliGemma instruction and use T5 as a supplementary semantic input. Therefore, when the native language input already generalizes reliably and controls the action, GSR retains it and uses T5 as a supplementary semantic source.

To sum up, as shown in Tab.~\ref{tab:main_results}, GSR raises Full Para success to 70.94\%, 49.12\%, and 75.59\% on VLA-Adapter, SmolVLA, and $\pi_{0.5}$, respectively. Xiaomi-Robotics-0 has the highest Full Para success among the reported systems at 76.0\%, narrowly above 75.59\% for GSR with $\pi_{0.5}$. GSR with $\pi_{0.5}$ achieves the highest reported PRIDE (70.4). We note a slight degradation in its canonical Goal SR (91.0\%) under the standard matched training schedule. This temporary drop stems from our architectural intervention, which requires re-initializing the action expert. For models equipped with larger, parameter-heavy action experts like $\pi_{0.5}$, this randomly initialized module naturally requires more optimization steps to fully converge. Fortunately, simply extending the training duration easily resolves this issue. When trained for twice the original number of steps, the canonical Goal SR of $\pi_{0.5}$ fully recovers to 96.0\%. We will further investigate more efficient initialization and training strategies for large action experts in future work.

These findings crystallize our core architectural principle. To achieve robust instruction following, the linguistic instruction must first be encoded \textit{independently} of the current dynamic image to extract pure, stable task semantics. Subsequently, these extracted semantics must be injected back into the model's computational pipeline to deeply interact with the visual scene and robot state. 

\begin{figure}[!t]
    \centering
    \includegraphics[width=1\linewidth]{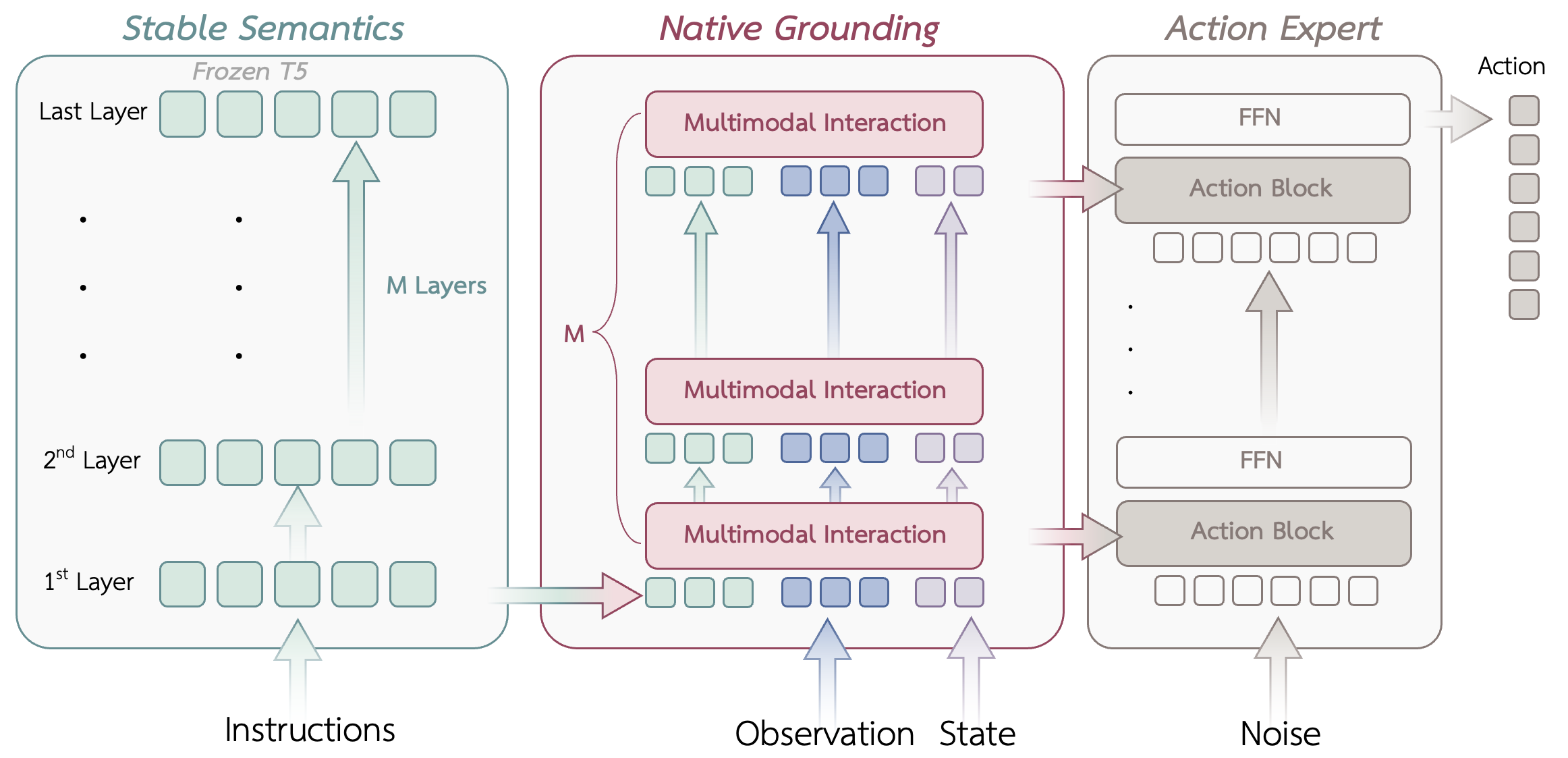}
    \caption{Overview of Grounded Semantic Re-binding.}
    \label{fig:gsr_overview}
    %\vspace{-1em}
\end{figure}

\subsection{General Guidelines for Language Following}
Building upon our previous empirical insights, Fig.~\ref{fig:gsr_overview} summarizes a general, architecture-aware guideline for implementing Grounded Semantic Re-binding (GSR). For any given VLA model, a frozen T5 encoder first maps the raw instruction to a pure semantic representation. T5 receives no image or robot state to ensure the extracted semantics remain entirely stable. Instead of appending a rigid external module, GSR identifies the specific computational stage where the native architecture performs its core multimodal fusion among task information, the visual scene, and the robot state. We inject the projected T5 semantics directly into this existing integration point. The exact timing of this integration naturally adapts to the specific architecture. For example, SmolVLA and $\pi_{0.5}$ mix the robot state jointly with vision and language early in their computation. In contrast, VLA-Adapter fuses vision and language first and incorporates the robot state later within its action-generating Bridge Attention component.

A critical step in this guideline is determining how to handle the model's native language input. We evaluate the original model using paired canonical and paraphrased instructions to test its native language reliability. If the native pathway is vulnerable to wording changes, as observed in VLA-Adapter and SmolVLA, we neutralize the original text input into a fixed neutral sentence. This forces the model to rely exclusively on the robust T5 semantics. Conversely, if the native pathway is already reliable, as seen in $\pi_{0.5}$, we retain the real instruction to provide complementary information. GSR therefore operates as a principled, architecture-aware methodology rather than a fixed structural patch.

%\newpage
\section{Further Analysis}
\label{sec:Further Analysis}

\subsection{Layerwise Control of Action Generation}

To understand exactly where different architectures process task semantics, we perform a layerwise intervention across all three models. For each canonical-paraphrase pair, we keep the visual observation, robot state, and action-generation noise strictly identical. We then systematically replace the internal features at one specific layer of the paraphrase run with the corresponding features from the canonical run. We evaluate the effectiveness of this intervention using a recovery metric. Let $B$ denote the number of paired predictions. We first compute the mean Euclidean action distance 
\begin{equation}
d_{\mathrm{act}}(x,y)=\frac{1}{B}\sum_{i=1}^{B}\left\lVert\operatorname{vec}(a_i^x)-\operatorname{vec}(a_i^y)\right\rVert_2.
\end{equation}
The recovery rate is then defined as $1-\frac{d_{\mathrm{act}}(\mathrm{patch},c)}{d_{\mathrm{act}}(p,c)}$, where $c$, $p$, and $\mathrm{patch}$ represent the canonical, paraphrase, and modified runs. This metric quantifies how much the substituted feature corrects the action back to the canonical trajectory.

\begin{figure}[!t]
    \centering
    \includegraphics[width=1 \linewidth]{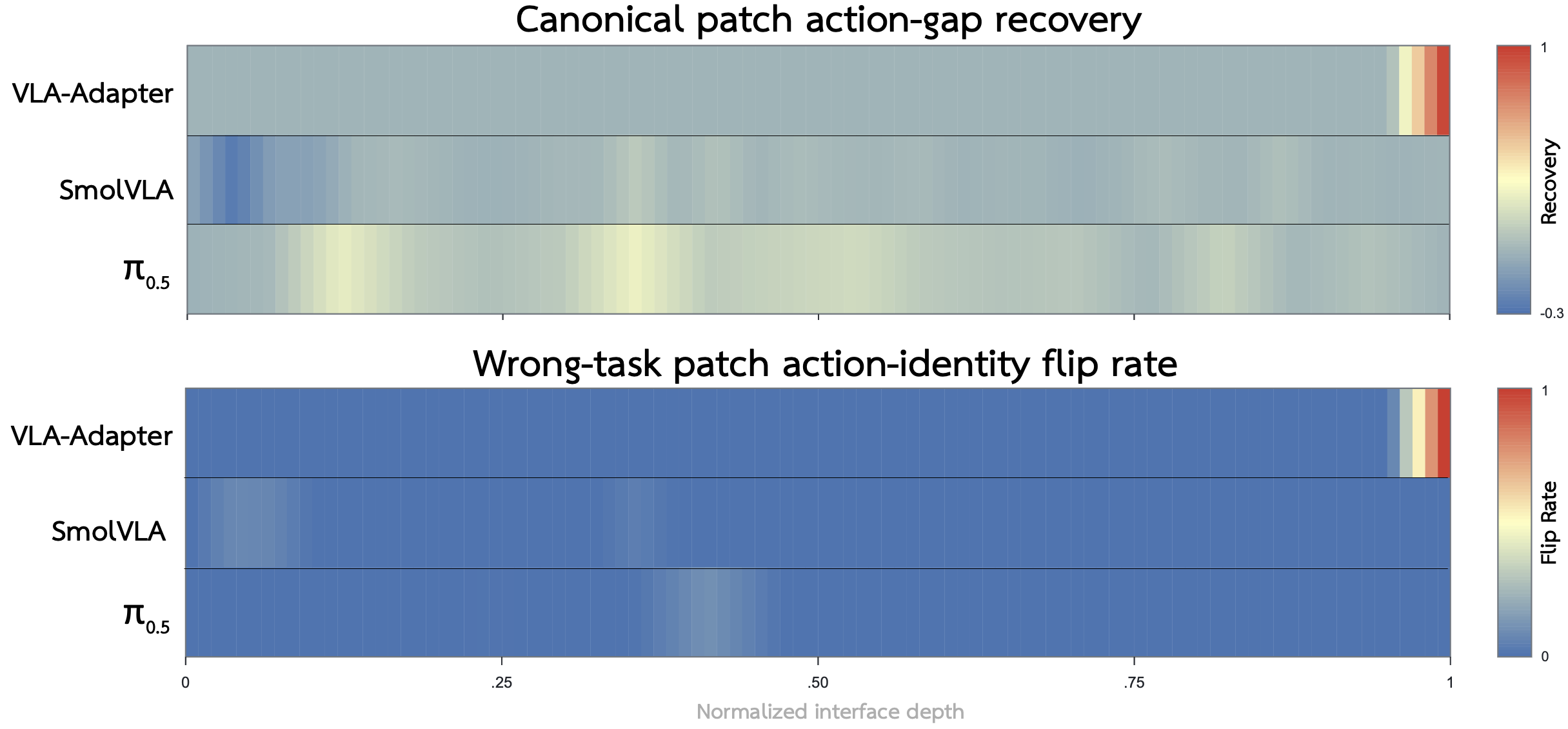}
    %\vspace{-1em}
    \caption{Layerwise interventions. The top panel shows action recovery after canonical-value replacement, while the bottom shows shifts toward an injected wrong task.}
    \label{fig:Fig4}
    %\vspace{-1em}
\end{figure}

As shown in Fig.~\ref{fig:Fig4}, the results reveal stark differences in how task information flows through different architectures. In VLA-Adapter, replacing features at the final Bridge-Attention block alone recovers 96.8\% of the action difference. To confirm this is not an arbitrary perturbation, we inject features from a completely different task and observe that the prediction directly shifts toward that wrong task. This proves the final block acts as a strict bottleneck for task information. In contrast, SmolVLA and $\pi_{0.5}$ exhibit no such single control point. Their best single-layer recovery rates are merely 10.5\% and 31.3\% respectively. This indicates that their task semantics are deeply entangled and processed in a distributed manner across multiple layers.

These structural differences stem from the architectural designs rather than simple training variations. For instance, SmolVLA freezes all 290 tensors in its text tower but still lacks a single correction point. VLA-Adapter, however, routes the language output into a separate Bridge-Attention policy where the final block directly precedes action generation. On the same diagnostic set of 100 paired canonical–paraphrase LIBERO-Goal episodes, training only this final block achieves 34\% paraphrase success, while training the final four blocks reaches 49\%. Both remain below the 75\% achieved when the complete policy is trained with T5 task conditions. These findings validate our core design philosophy. Because different models fuse task information at different stages, a fixed structural patch will inevitably fail.

\subsection{Language-Source and Route Ablations}
\begin{wraptable}{r}{0.50\textwidth}
    \vspace{0pt}
    \centering
    \small
    \setlength{\tabcolsep}{3.5pt}
    \renewcommand{\arraystretch}{1.08}
    \resizebox{\linewidth}{!}{%
    \begin{tabular}{@{}lccccc@{}}
        \toprule
        Model & C/C & W/C & C/W & Native Drop & T5 Drop \\
        \midrule
        $\pi_{0.5}$ + T5 & \textbf{75.59} & 2.88 & 74.0 & 72.7 & 1.6 \\
        VLA-Adapter + T5 & \textbf{47.31} & 5.11 & 44.0 & 42.2 & 3.3 \\
        SmolVLA + T5 & \textbf{13.49} & 2.25 & 3.23 & 11.24 & 10.26 \\
        \bottomrule
    \end{tabular}%
    }
    \caption{Language-route authority. Native Drop and T5 Drop are measured relative to C/C.}
    \label{tab:language_route}
    \vspace{-0.5\baselineskip}
\end{wraptable}

We then verify that our performance gains come specifically from the pure semantics of T5 rather than just the addition of trainable parameters. In our VLA-Adapter diagnostic tests, the native policy achieves only 46.82\% success on paraphrased instructions. Simply adding extra trainable parameters without a language model, or replacing T5 with Qwen-VL, yields the exact same 46.82\% success rate. However, injecting our frozen T5 semantics boosts paraphrase success to 70.94\%. This confirms that merely increasing model capacity cannot solve the grounding failure. The improvement relies entirely on the stable, vision-free semantic features provided by T5.

Next, we investigate a critical question regarding internal control. When a model receives both its native language input and the injected T5 semantics, which signal actually drives the robot's behavior? To answer this, we design a conflict test across all three models on the full LIBERO-Para distribution. As shown in Tab.~\ref{tab:language_route}, we intentionally feed mismatched instructions to the two pathways. The C/C setting gives both inputs the correct paraphrased instruction. The W/C setting gives a wrong instruction to the native encoder while giving the correct one to T5. Conversely, the C/W setting gives the correct instruction to the native encoder and a wrong one to T5. This setup clearly reveals which pathway dominates the final action generation.

The results perfectly justify our architecture-aware guidelines. For VLA-Adapter, feeding a wrong instruction to its native pathway causes the success rate to plummet from 47.31\% to a mere 5.11\%. Conversely, giving a wrong instruction to T5 only drops performance to 44.0\%. This reveals a dangerous flaw where the model stubbornly obeys its native Qwen pathway even though that pathway generalizes poorly to paraphrases. SmolVLA suffers equally in both mismatch scenarios, indicating that the two active pathways severely interfere with each other. In contrast, $\pi_{0.5}$ relies heavily on its native PaliGemma pathway, but unlike Qwen, PaliGemma is inherently robust to wording changes. 

These findings prove that an unreliable native encoder can still hijack the model's control. Therefore, as dictated by our GSR methodology, we must neutralize the original text input for vulnerable models like VLA-Adapter and SmolVLA to force reliance on T5, while retaining it for robust models like $\pi_{0.5}$ to leverage complementary information.

\subsection{Real-Robot Experiments}

\begin{wrapfigure}{r}{0.48\textwidth}
    \vspace{-0.8\baselineskip}
    \centering
    \includegraphics[width=\linewidth]{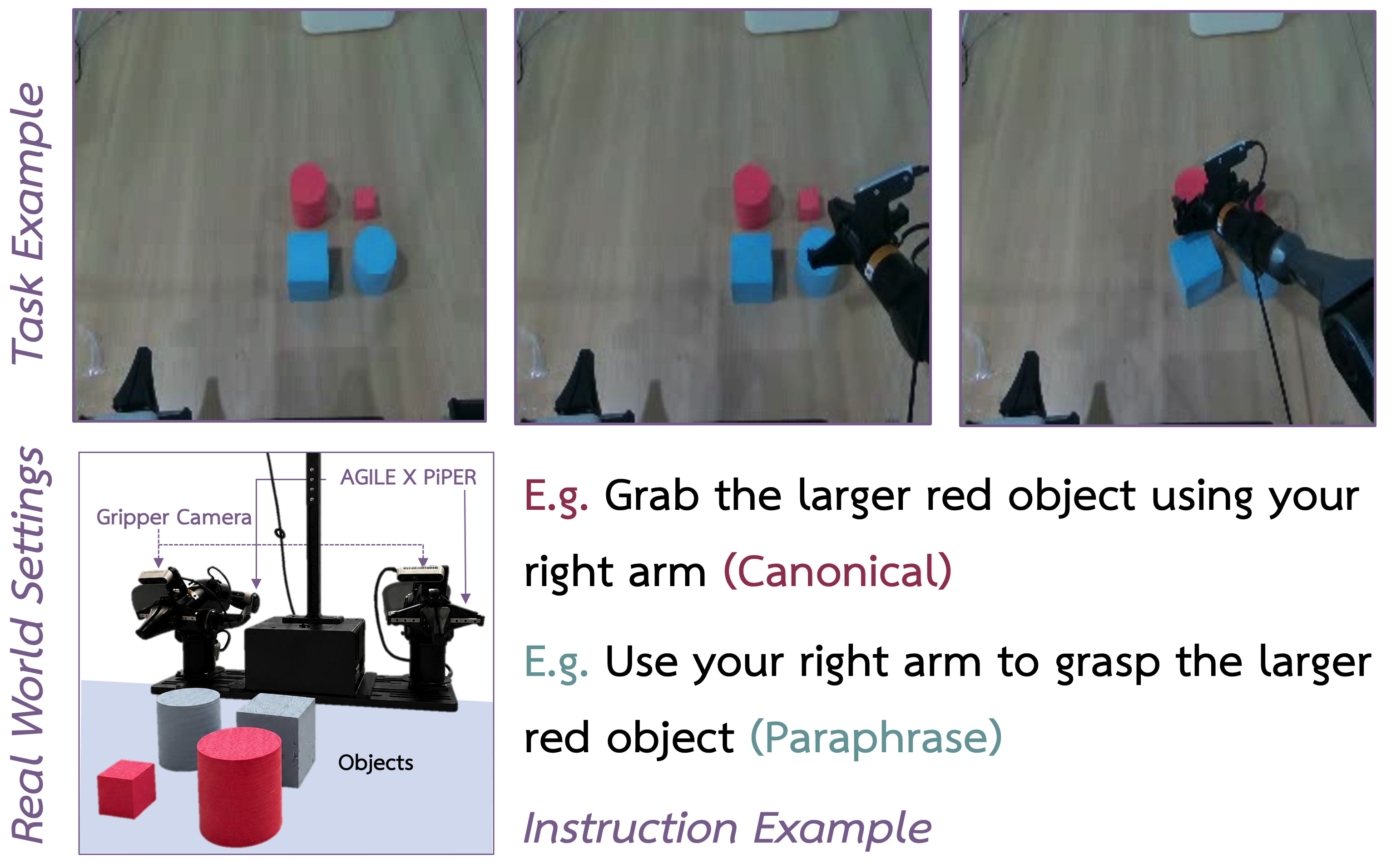}
    \caption{Real-world system and task examples.}
    \label{fig:real}
    \vspace{-0.5\baselineskip}
\end{wrapfigure}

In a real-robot pilot, as shown in Fig.~\ref{fig:real}, we evaluate VLA-Adapter on four pickup and two stacking tasks using the same robot, camera setup, and initial arrangement of four objects. The visual scene is therefore shared across tasks, while the natural-language instruction and the corresponding success criterion specify which object or relation is requested. Each task is evaluated with its canonical instruction and an unseen paraphrase, and the pretrained VLM remains frozen during training. The complete task definitions, instructions, and evaluation protocol are provided in the supplementary material. Because the shared scene does not reveal which task should be performed, the policy must distinguish the tasks from language. The native VLA-Adapter fails to use the frozen VLM features for this distinction and collapses to the same task-independent motion, obtaining 0\% success under both canonical and paraphrased instructions. GSR instead achieves 50\% and 40\% success, respectively.

%\input{sections/5_conclusion}
%\clearpage
\section{Discussion: Natively Decoupled Alternative}
\label{sec:discu}

\begin{wrapfigure}{R}{0.48\textwidth}
    \vspace{0pt}
    \centering
    \captionsetup{font=footnotesize}
    \includegraphics[width=0.88\linewidth]{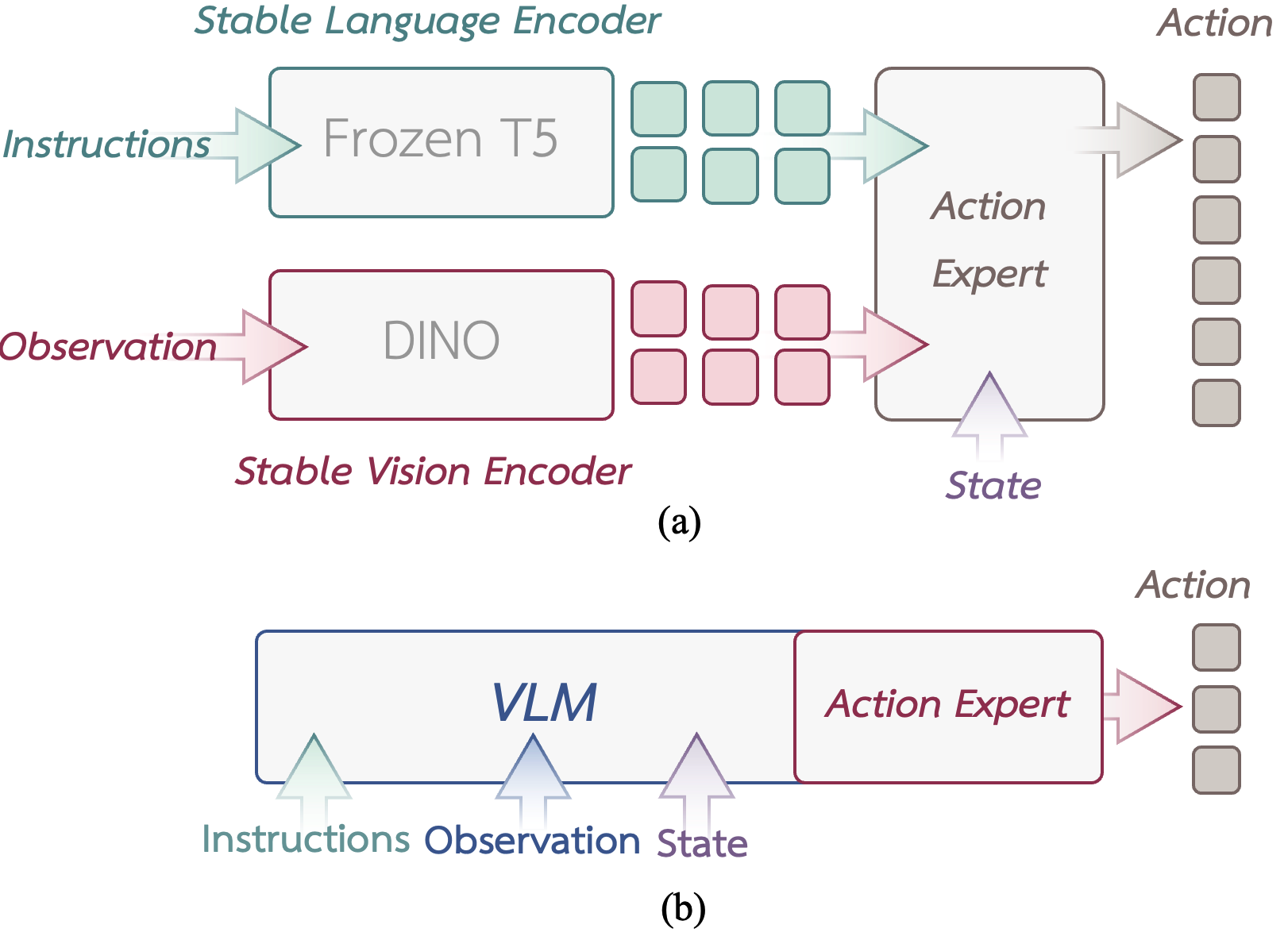}
    \caption{ParaVLA and conventional VLA architectures. \textbf{(a)} ParaVLA; \textbf{(b)} classic VLA.}
    \label{fig:ParaVLA}
    \vspace{-1\baselineskip}
\end{wrapfigure}

Our analyses show that joint VLM encoding makes task representations sensitive to wording, whereas architecture-aware injection of frozen T5 semantics improves robustness. This raises whether language stability can instead be achieved by separating language and vision throughout representation extraction. We therefore design ParaVLA, a decoupled baseline that removes joint vision-language encoding, processes instructions with frozen T5 and images with DINOv2~\cite{dinov2}, and combines them only in the final action expert(Fig.~\ref{fig:ParaVLA}).

As shown in Tab.~\ref{tab:goal_paraphrase_comparison}, experiments confirm the viability and extreme robustness of this pure decoupled approach. With the visual observation and action noise fixed, ParaVLA responds strongly to actual task changes but remains highly stable against wording variations. Closed-loop evaluations further validate this design. ParaVLA achieves 92.0\% canonical and 91.0\% paraphrase success with 0.33B parameters active per control step. T5 runs once per task, and its cached output adds no per-step inference cost. When evaluating the combined reliability across both canonical and paraphrased instructions, ParaVLA achieves the best performance among current models at this scale. To verify that standard VLMs inherently pollute language representations and that a stable semantic source is strictly required, we replace T5 with a SmolVLM decoder within this exact same architecture. While canonical success remains at 85.0\%, paraphrase success collapses to 41.0\%. This proves that completely discarding the VLM in favor of a decoupled T5 and DINO architecture effectively solves the language pollution problem from the ground up.

\begin{table}[t]
\centering
\small
\setlength{\tabcolsep}{3pt}
\renewcommand{\arraystretch}{1.08}
\begin{tabular}{llrrrrr}
\toprule
 & Model & Canon.$\uparrow$ & Para.$\uparrow$ & Drop$\downarrow$ & Full Para$\uparrow$ & PRIDE$\uparrow$ \\
\midrule
\textcircled{1} & VLA-Adapter(0.51B) & 94 & 56 & 38 & 46.82 & 36.7 \\
\textcircled{2} & SmolVLA(0.45B) & 92 & 7 & 85 & 4.47 & 2.6 \\
\textcircled{3} & $\pi_{0.5}$(3.6B) & 96 & 90 & $6$ & 73.60 & -- \\
\textcircled{4} & \textbf{ParaVLA}(0.33B) & \textbf{92} & \textbf{91} & \textbf{1} & \textbf{72.51} & \textbf{66.9} \\
\bottomrule
\end{tabular}
\caption{Canonical and paraphrased LIBERO-Goal success rates, together with Full LIBERO-Para results. Drop is computed as Canon. minus Para. Parameter counts include frozen and trainable components.}
\label{tab:goal_paraphrase_comparison}
\end{table}

\textbf{Limitation and Future Research.} Despite its exceptional robustness to language variations, this decoupled framework encounters a critical bottleneck regarding scalability. When we attempt to increase the parameter scale of the visual foundation model, such as upgrading to a larger DINO variant, the system fails to exhibit the strong scaling capabilities typically observed in VLMs (\textcircled{4}). Consequently, while natively decoupled models offer a perfect solution for semantic stability, achieving VLM-level scaling within this pure paradigm remains a significant challenge. Addressing this scaling bottleneck in decoupled architectures represents an important direction for our future exploration.

%Due to space constraints, the section of related work, experimental details, and additional results are deferred to the appendix. Code and models will be open-sourced to facilitate future research.

%\newpage
\section{Related Work}
\label{sec:related}
\subsection{Vision–language–action models} 
Vision–language–action (VLA) models build on language-conditioned imitation learning and behavioral transformers to learn generalist policies from diverse tasks, datasets, and robot embodiments. Their designs span autoregressive, diffusion, and flow-based action generation, together with spatial, cross-embodiment, and compact architectures~\cite{rt1,visionlanguagefoundationmodelseffective,rt2,openvla,cogact,rdt,dexvla,pi0,pi05,xvla,smolvla,vla-adapter,tinyvla} . A parallel line of work analyzes what task and state information these policies encode and whether intervening on their internal representations changes behavior, and how language-action alignment can be verified or adjusted~\cite{grant2026featurescreatedequalmechanistic,swann2026sparseautoencodersrevealinterpretable,dong2026doeslanguagemattermultilingual,rosa,buurmeijer2026observingcontrollingfeaturesvisionlanguageaction,mitra2025mechanisticfinetuningvisionlanguageactionmodels,wu2026saysteeringvisionlanguageactionmodels,disc,kwok2026scalingverificationeffectivescaling,roboalign}. Yet standard task suites often rely on canonical instructions, while recent robustness studies show that high task success can coexist with memorization, distribution-shift failures, and weak instruction use~\cite{libero,liberopara,liberoplus,liberopro}. We study how language information is preserved and used during VLA action generation.

\subsection{Language generalization and following}
Language generalization in robot learning includes following unseen expressions, grounding open-world references, and executing new compositions of familiar skills and entities~\cite{lynch2022interactivelanguagetalkingrobots,myers2023goalrepresentationsinstructionfollowing,stone2023openworldobjectmanipulationusing,bcz,ocean,vlmbench,programmaticallygrounded,myers2025temporalrepresentationalignmentsuccessor} . Recent benchmarks further isolate failures under paraphrases, irrelevant context, multilingual instructions, scene perturbations, and controls for memorization~\cite{liberopara,applesofa,dong2026doeslanguagemattermultilingual,liberopro,liberoplus,mcolosseumv2}. Existing methods improve robustness through instruction relabeling, counterfactual supervision, semantic or visual augmentation, representation alignment, consistency training, and inference-time language steering~\cite{glossop2026castcounterfactuallabelsimprove,chen2024semanticallycontrollableaugmentationsgeneralizable,stablelanguageguidance,rovla,jeong2026learningsayvlaharmless}. These approaches broaden linguistic coverage or introduce additional robustness supervision. Our GSR takes a complementary route. It uses only canonical demonstrations, extracts task semantics independently of the current observation, grounds them through the model's native multimodal computation, and relearns their mapping to robot actions.

\section{Conclusion}
\label{sec:conclu}

Across three representative VLAs, we observe that while paraphrased instructions cause execution failures, task-specific structures remain detectable in predicted actions. Controlled experiments further reveal that architectures differ significantly in how task inputs influence action generation. To address this, GSR extracts task information independently of current observations and grounds it via each model's scene computation. Trained solely on canonical demonstrations, GSR improves success on fully paraphrased instructions across all models, achieving a 70.4 PRIDE score.

\appendix
\section*{Appendix}

\section{Benchmark and Evaluation Protocol}

\subsection{LIBERO-Goal Training Data}

The LeRobot LIBERO~\cite{libero} dataset slice used in our experiments consists of episodes 379--806, inclusive, totaling 428 episodes and 52,042 transitions. It is the LIBERO-Goal subset and contains two image streams, robot state, and 7-DoF actions. The VLA-Adapter~\cite{vla-adapter} experiments use the corresponding released VLA-Adapter/LIBERO-Goal-Pro initialization and canonical Goal RLDS demonstrations. All final policies are trained only on canonical demonstrations; no LIBERO-Para~\cite{liberopara} instruction or manual diagnostic paraphrase is used for training.

\subsection{Full LIBERO-Para}

All configurations compared within the same model use identical episode keys and instruction metadata. The paired unit is the episode, not an aggregate task score.

\subsection{Goal-100 and Manual-Paraphrase Diagnostics}

\begin{wraptable}{r}{0.48\textwidth}
\vspace{-1.2\baselineskip}
\centering
\footnotesize
\begin{tabular}{@{}lrl@{}}
\toprule
Axis & Episodes & Description \\
\midrule
Act & 870 & Action wording changes \\
Obj & 259 & Object-reference wording changes \\
Comp & 2,963 & Combined action and object changes \\
\textbf{Total} & \textbf{4,092} & Official full benchmark \\
\bottomrule
\end{tabular}
\caption{Composition of the full LIBERO-Para evaluation.}
\label{tab:supp-01}
\vspace{-0.5\baselineskip}
\end{wraptable}

Goal-100 uses the ten LIBERO-Goal tasks and ten fixed official initial states per task. For each task, the same observation and initial state are paired with the canonical instruction, one fixed manual paraphrase, and a cyclic wrong-task instruction from the next task.

Manual-paraphrase text is used only for evaluation and diagnostics.

\begin{table*}[t]
\centering
\small
\begin{tabularx}{\textwidth}{rXXX}
\toprule
Task & Canonical instruction & Fixed manual paraphrase & Cyclic wrong-task instruction \\
\midrule
00 & open the middle drawer of the cabinet & slide open the cabinet's center drawer & put the bowl on the stove \\
01 & put the bowl on the stove & set the bowl down onto the stovetop & put the wine bottle on top of the cabinet \\
02 & put the wine bottle on top of the cabinet & place the bottle of wine atop the cabinet & open the top drawer and put the bowl inside \\
03 & open the top drawer and put the bowl inside & pull the upper drawer open and drop the bowl in it & put the bowl on top of the cabinet \\
04 & put the bowl on top of the cabinet & rest the bowl up on the cabinet's top surface & push the plate to the front of the stove \\
05 & push the plate to the front of the stove & nudge the plate forward to the stove's front edge & put the cream cheese in the bowl \\
06 & put the cream cheese in the bowl & drop the cream cheese block into the bowl & turn on the stove \\
07 & turn on the stove & switch the stove on & put the bowl on the plate \\
08 & put the bowl on the plate & lay the bowl down onto the plate & put the wine bottle on the rack \\
09 & put the wine bottle on the rack & place the bottle of wine onto the rack & open the middle drawer of the cabinet \\
\bottomrule
\end{tabularx}
\caption{Canonical, manually paraphrased, and cyclic wrong-task instructions used in the Goal-100 diagnostic.}
\label{tab:supp-02}
\end{table*}

For the standard Goal-100 wrong-task condition, task $i$ receives the canonical instruction of task $(i+1)\bmod 10$. The dual-language route-conflict experiment is evaluated separately on the full LIBERO-Para benchmark, as detailed below.

\subsection{Pairing and Stochastic Control}

For a paired comparison, we hold fixed the BDDL task instance and initial state, environment stabilization and success predicate, observation and robot state for in-silico probes, episode and policy seed, and the flow noise, timestep, and sampling seed for flow models.

The fixed-observation traces use a fixed flow timestep $t=0.5$ and seed 35. In the locked full-4092 comparison, $\pi_{0.5}$ executes five actions per replanning step, VLA-Adapter executes eight, and SmolVLA executes one. These horizons are model-specific validated contracts and are not cross-model hyperparameters.

\subsection{Metrics}

We report success rate and the official PRIDE metric. PRIDE assigns each paraphrase a deviation weight $\mathrm{PD}_i(\alpha)$ derived from keyword and structural similarity; we use the official default $\alpha=0.5$.

\begin{equation}
\label{eq:pride}
\begin{aligned}
\mathrm{PRIDE}(\alpha)
&=100\,
\frac{\sum_i y_i\mathrm{PD}_i(\alpha)}
{\sum_i \mathrm{PD}_i(\alpha)}.
\end{aligned}
\end{equation}

We report exact two-sided McNemar tests from paired success/failure outcomes. Where available, task-stratified bootstrap 95\% confidence intervals resample tasks rather than treating all paraphrases as independent linguistic units. Wilson intervals are used for unpaired binomial summaries.

\section{Architecture and Implementation Details}

\subsection{Common Stable Language Source}

The stable source is frozen T5-large~\cite{t5}. We use its token-level final encoder hidden states and mask padding. T5 receives the real task instruction. T5 parameters remain frozen, while architecture-specific projection and grounding modules are trainable.

The neutral native prompt is exactly ``perform the task.''

The neutral prompt removes task identity from the fragile native text route but does not remove images, robot state, or the native multimodal scene computation.

\subsection{VLA-Adapter}

\subsubsection{Computation}

The main Qwen~\cite{qwen} path receives images and the neutral prompt. The real instruction is encoded by frozen T5-large. Token-level T5 features are projected to the action-head width and supplied as a sidecar K/V source to each MLPResNetBlock\_Pro. The sidecar has its own attention softmax and per-block gate; its output is combined with the block's native action update. The action head is trained from random initialization with canonical L1 imitation.

The key distinction from a generic concatenation is that Qwen visual/task conditioning and T5 task semantics remain separately addressable inside the action policy.

The VLA-Adapter representation audit separates three sources: visual-prefix keys and values, seven Qwen instruction-conditioned tokens plus proprioception, and semantic sidecar K/V formed by projected T5 tokens with an independent softmax.

VLA-Adapter GSR uses 24 action blocks, an 8-step, 7-dimensional action chunk, a frozen Qwen backbone, frozen T5-large, and a fresh action head. The final output is deterministic L1 regression.

\subsection{SmolVLA}

The SmolVLA~\cite{smolvla} GSR configuration does not append T5 as a late foreign action sidecar. SmolVLM~\cite{smolvlm} receives images and the neutral prompt, producing native language slots in its multimodal prefix. Frozen T5 encodes the real instruction and is projected from 1024 to 960 dimensions.

The native language slots query projected T5 tokens through an 8-head grounding attention.

\begin{equation}
\label{eq:smolvla-grounding}
\tilde h_{\text{task}}
=h_{\text{native-task}}+g\,\mathrm{MHA}\bigl(\mathrm{LN}(h_{\text{native-task}}),
\mathrm{LN}(P_{\mathrm{T5}}),\mathrm{LN}(P_{\mathrm{T5}})\bigr).
\end{equation}

The grounded task slots then continue through SmolVLM's native multimodal processing and condition the flow-matching expert. Thus the stable semantics see the scene through the original architecture before action generation.

The native-slot grounder uses eight attention heads, its gate is initialized to 1.0, and the native instruction is replaced by the fixed prompt ``perform the task.''

\subsection{$\pi_{0.5}$}

Because $\pi_{0.5}$ was exposed to large and diverse datasets during pretraining, its native PaliGemma~\cite{paligemma} route already provides strong semantic control. We therefore preserve the real instruction in $\pi_{0.5}$ GSR and use frozen T5-large as a supplementary grounded signal. Following the same grounding strategy as SmolVLA, projected T5 tokens are grounded at dynamically detected PaliGemma task-token positions before entering the native multimodal stack.

The task mask begins at the first instruction token and ends before the final ``State:'' delimiter. It never overwrites image tokens, BOS/EOS, padding, state tokens, or action-template tokens. A fresh 300M action expert is trained with the original flow-matching objective while the PaliGemma/VLM backbone remains frozen.

\section{Training Configurations}

\subsection{Locked Training Recipes}

All configurations use canonical demonstration language only. Frozen T5 is an encoder, not a language-training objective.

\begin{center}
\centering
\footnotesize
\begin{tabularx}{\columnwidth}{@{}p{0.34\columnwidth}X@{}}
\toprule
Configuration & Value \\
\midrule
GPU model & $8\times$ NVIDIA RTX 4090 \\
Batch size & 8 per GPU / 64 global \\
Steps & 50,000 \\
Optimizer & AdamW, peak LR $10^{-4}$, weight decay $5\times10^{-4}$ \\
\bottomrule
\end{tabularx}
\captionof{table}{Reference training configuration for ParaVLA.}
\label{tab:supp-06}
\end{center}

\begin{center}
\centering
\footnotesize
\begin{tabularx}{\columnwidth}{@{}p{0.34\columnwidth}X@{}}
\toprule
Configuration & Value \\
\midrule
GPU model & $8\times$ NVIDIA RTX 4090 \\
Batch size & 8 per GPU / 64 global \\
Steps & 50,000 \\
Optimizer & AdamW, peak LR $5\times10^{-4}$, minimum LR $5\times10^{-5}$, weight decay $10^{-4}$ \\
\bottomrule
\end{tabularx}
\captionof{table}{Training configuration for VLA-Adapter GSR.}
\label{tab:supp-07}
\end{center}

\begin{center}
\centering
\footnotesize
\begin{tabularx}{\columnwidth}{@{}p{0.34\columnwidth}X@{}}
\toprule
Configuration & Value \\
\midrule
GPU model & $8\times$ NVIDIA RTX 4090 \\
Batch size & 16 per GPU / 128 global \\
Steps & 25,000 \\
Optimizer & AdamW, LR $2\times10^{-4}$, weight decay $10^{-10}$ \\
\bottomrule
\end{tabularx}
\captionof{table}{Training configuration for SmolVLA GSR.}
\label{tab:supp-08}
\end{center}

All three $\pi_{0.5}$ configurations initialize the VLM from the LIBERO-finetuned $\pi_{0.5}$ checkpoint and reinitialize the gemma\_300m action expert. The no-T5 control uses the real native instruction and no additional language route. Both GSR configurations preserve that instruction and ground frozen T5 tokens into the native task slots.

\begin{center}
\centering
\footnotesize
\begin{tabularx}{\columnwidth}{@{}p{0.34\columnwidth}X@{}}
\toprule
Configuration & Value \\
\midrule
GPU model & $8\times$ NVIDIA A800 \\
Batch size & 64 per GPU / 512 global \\
Steps & Native control: 6,250; GSR matched: 6,250; GSR$^{\ast}$ extended: 12,500 \\
Optimizer & Matched-control recipe; GSR$^{\ast}$ uses the same recipe with extended steps \\
\bottomrule
\end{tabularx}
\captionof{table}{Matched-control, matched GSR, and extended GSR$^{\ast}$ training configurations for $\pi_{0.5}$.}
\label{tab:supp-09}
\end{center}

Matched GSR uses the same update count and sample exposure as the Native matched control. GSR$^{\ast}$ keeps the same architecture and training recipe but doubles the number of updates and total sample exposure. No paraphrase distillation, interface matching, soft-prefix loss, motor distillation, or K/V matching objective was enabled.

\subsection{Implementation Notes}

The retrofitted GSR policies compute frozen-T5 features online, whereas ParaVLA evaluates T5 once per task and caches its token features.

\FloatBarrier

\section{Diagnostic Definitions and Interventions}

\subsection{Fixed-Observation Audit}

The fixed-observation audit uses 100 real observations, comprising ten first frames per Goal task. Each observation is reused with all ten canonical, paraphrase, and cyclic wrong-task instructions. Means and top principal components for anisotropy control are fit from an independent four-suite canonical corpus, not from the ten test pairs.

For the $n$-th test case, let $y_n$ denote its ground-truth task label, $a_n^p$ the flattened action chunk generated from the paraphrased instruction, and $a_{n,k}^c$ the action chunk generated from the canonical instruction of task $k$ under the same observation. We use the centered-and-normalized distance, Retrieval@1, and semantic-retention score $R$ defined in the main paper. The corresponding wrong-task to paraphrase distance ratio is reported as $\Delta_{\mathrm{wrong}}/\Delta_{\mathrm{para}}$.

The same definitions are applied to vectorized hidden-state captures using their corresponding canonical-corpus mean. Chance Retrieval@1 is $1/N=0.1$. Action chunks are flattened after each model's normal decoding. Cross-model claims use scale-free retrieval, retention, ratios, cosine, or relative RMSE; raw distances are not compared across architectures.

\subsection{Layerwise Capture}

The offline capture retains T5 layers, text projection, and pre/text-attention/post states from ParaVLA's eight action blocks; the Qwen task condition and pre/KV/residual/post states from VLA-Adapter's 24 action blocks; task-language K/V and velocity/action states across SmolVLA's 32 VLM-prefix/expert layers; and task-span K/V, expert-suffix, and velocity/action states across $\pi_{0.5}$'s 18 PaliGemma layers.

Capture hooks are read-only by default. Disabling capture reproduces the checkpoint output exactly or within declared BF16 tolerance.

\begin{figure*}[t]
\centering
\includegraphics[width=\textwidth]{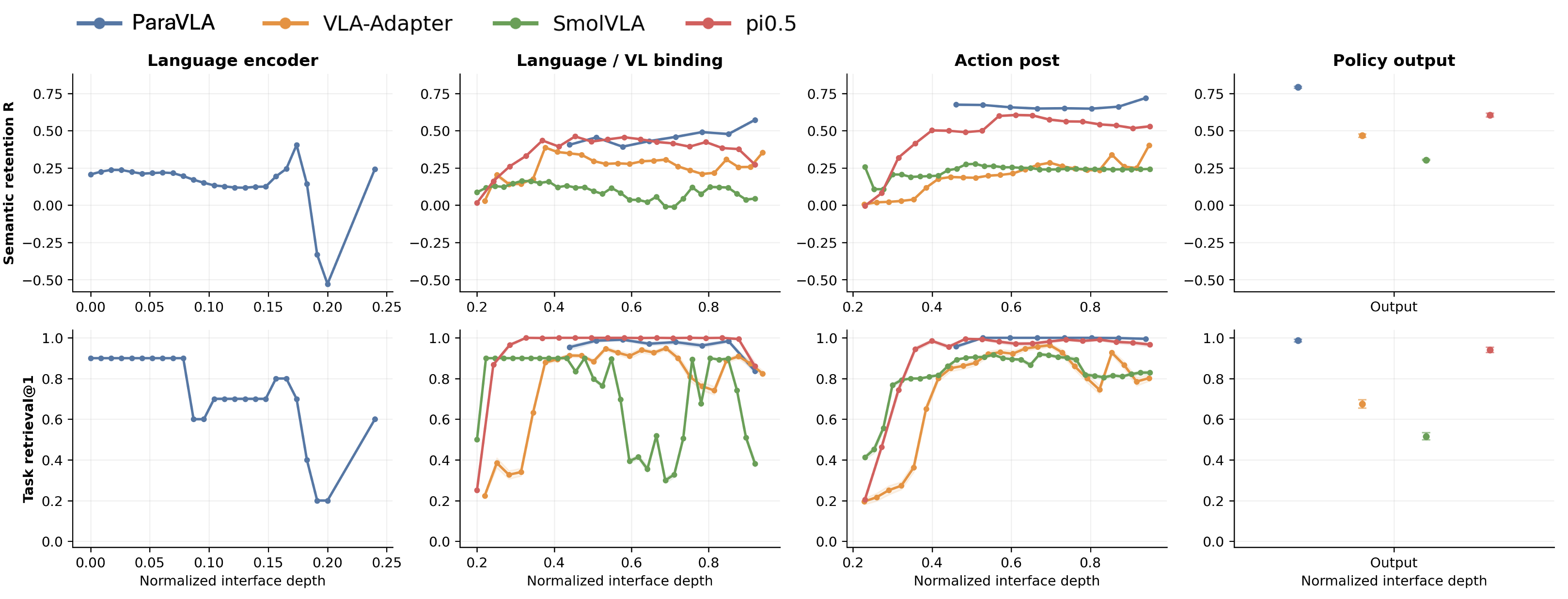}
\caption{Stage-aligned instruction diagnostics across ParaVLA, VLA-Adapter, SmolVLA, and $\pi_{0.5}$. Interfaces are shown in forward order within the language encoder, language/vision--language binding, action-processing, and policy-output stages. The top row reports semantic retention $R$, which contrasts paraphrase and wrong-task distances; the bottom row reports Retrieval@1. Curves connect interfaces only within the same model and stage, while the isolated points show the final policy outputs. Higher values indicate stronger preservation of task identity.}
\label{fig:four-model-depth-trace}
\end{figure*}

Fig.~\ref{fig:four-model-depth-trace} provides the stage-aligned view of these diagnostics. Because the four architectures expose different numbers and types of interfaces, normalized depth indicates ordering within an architecture-specific stage; equal horizontal positions do not imply homologous layers across models.

\subsection{Activation Patching}

We use the mean Euclidean action distance $d_{\mathrm{act}}$ and recovery metric defined in the main paper. Canonical, paraphrase, and layer-$l$ patched runs are denoted by $c$, $p$, and $\mathrm{patch}_l$, respectively.

Canonical-to-paraphrase patches test recoverability. Wrong-task-to-canonical patches test task control. Identity patches must leave output unchanged.

We patch architecture-specific channels. ParaVLA uses the dedicated text-attention projected K/V; VLA-Adapter uses the first seven instruction-conditioned tokens or their attention residual while preserving proprioceptive and visual tokens; SmolVLA uses native prefix language-token K/V; and $\pi_{0.5}$ uses PaliGemma task-span native K/V.

A predetermined diagnostic criterion for the stronger claim of a \emph{universal semantic breakpoint}, applicable across architectures, required all three conditions.

\begin{enumerate}
\item a retention drop of at least 0.10 with a task-stratified bootstrap lower confidence bound above zero;
\item exact action-gap recovery at least 0.50;
\item significant wrong-task action-identity control.
\end{enumerate}

No model satisfied all three conditions for a universal cross-architecture semantic breakpoint. This stronger criterion is distinct from the model-specific action-control bottleneck identified in the main paper. In VLA-Adapter, the final Bridge-Attention block (block 23) recovers 96.8\% of the canonical--paraphrase action gap when patched with the canonical feature, while a wrong-task feature shifts the prediction toward the injected task. We therefore describe this final block, consistently with the main paper, as a strict model-specific bottleneck through which task information controls action generation; we do not claim that an equivalent single bottleneck exists in every architecture.

\subsection{Route-Conflict Interventions}

\begin{figure}[t]
\centering
\includegraphics[width=\columnwidth]{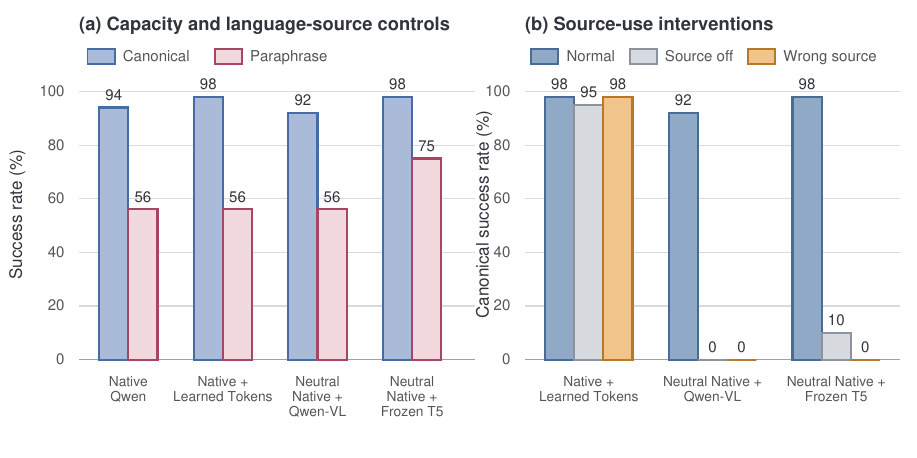}
\caption{Language-source factorization for VLA-Adapter on the Goal-100 diagnostic. (a) Capacity controls show that only the frozen-T5 route improves paraphrase success, from 56\% to 75\%. (b) Removing or corrupting the Qwen--VL or frozen-T5 task source sharply degrades canonical success, while the learned-token control remains largely unchanged.}
\label{fig:language-source-factorization}
\end{figure}

For models with native and T5 routes, the evaluation covers five conditions: correct native/correct T5, wrong native/correct T5, correct native/wrong T5, both routes wrong, and neutral native/T5 off.

The route-conflict experiment uses all 4,092 Full LIBERO-Para episodes. C/C gives both routes the correct paraphrased instruction. W/C gives the native route a cyclically mismatched paraphrase while keeping the correct instruction for T5, and C/W applies the same mismatch only to T5. Correct and mismatched instructions use the same rewrite type. The intervention changes only text, not the episode, image, robot state, initialization, or model weights.

Fig.~\ref{fig:language-source-factorization} further separates additional conditioning capacity from reliance on a task-informative language source in VLA-Adapter.

\subsection{Wording-Subspace Controls}

Using five-fold task-disjoint cross-validation, we estimate 32 wording directions on eight tasks and apply them to the two held-out tasks, comparing form-subspace, equal-rank/equal-norm random, and task-subspace deletion. Removing the wording directions reduces the canonical-paraphrase action gap from 0.4361 to 0.2282, whereas random deletion leaves it at 0.4386. In 20 closed-loop episodes, wording-direction removal increases success from 55\% to 90\%.

\section{Real-Robot Protocol}

We conduct a small-scale real-robot pilot on six tasks that share the same workspace. The first four tasks require picking up a specified object, and the last two require stacking one specified object on another. All commands use the right arm.

\subsection{Hardware and Workspace}

The platform is an AgileX PiPER dual-arm robot equipped with three cameras. The head camera is fixed above the space between the two arms. Two wrist cameras are mounted on the robot arms. The six tasks use the same initial arrangement of four objects, so the requested behavior must be identified from the instruction.

The observation contains a 14-dimensional proprioceptive state and three RGB streams from the head, left-wrist, and right-wrist cameras. Each camera records 480-by-640 images. Images are converted from BGR to RGB and resized to $224\times224$ pixels by the RLDS pipeline before the vision-language processor.

The policy predicts 14-dimensional absolute joint-space targets ordered as six left-arm joints, the left gripper, six right-arm joints, and the right gripper. Joint angles use PiPER encoder units in millidegrees, and gripper travel uses 0.001 mm units. Actions are bound-normalized to $[-1,1]$ during training and denormalized before execution. Although every task instruction specifies the right arm, the policy outputs targets for both arms. The predicted action chunk contains 25 steps.

\subsection{Demonstrations and Collection Timing}

Demonstrations were collected by dual-arm teleoperation. The complete real-robot training set contains eight tasks with 40 episodes per task, giving 320 episodes and 27,648 action transitions. Training uses all eight tasks, while the real-robot evaluation uses the six tasks described above. The policy is trained with the full 14-dimensional action even though the left arm remains close to its home configuration in these right-arm tasks. The saved HDF5 trajectories run at 15 Hz.

Tab.~\ref{tab:piper-training} summarizes additional optimization details for the GSR checkpoint used for real-robot deployment.

\begin{center}
\footnotesize
\begin{tabularx}{\columnwidth}{@{}p{0.34\columnwidth}X@{}}
\toprule
Configuration & Value \\
\midrule
GPU model & $8\times$ NVIDIA A800 \\
Batch size & 8 per GPU / 64 global \\
Steps & 60,000 \\
Optimizer & AdamW, peak LR $5\times10^{-4}$, final LR $5\times10^{-5}$, weight decay $10^{-4}$ \\
\bottomrule
\end{tabularx}
\captionof{table}{Additional optimization details for the deployed GSR real-robot policy.}
\label{tab:piper-training}
\end{center}

The Native VLA-Adapter control uses the same trainable components, batch size, optimizer, and training schedule. It differs only in its language route: the VLM receives the original task instruction through the standard native interface, with no T5 sidecar.

\subsection{Tasks and Instructions}

The canonical instructions and unseen OOD paraphrases are listed in Tab.~\ref{tab:supp-24}.

\subsection{Evaluation}

Before each rollout, both robot arms are returned to their zero (home) configuration. For safety, the real-robot closed-loop evaluation runs at 1 Hz to limit rapid unintended motion. An episode terminates immediately when the task succeeds; otherwise, it runs for approximately 200 control steps, corresponding to about 200 seconds.

The repeated evaluation uses five trials per task and instruction condition, giving 30 trials for each policy--instruction route. Per-task outcomes are reported in Tab.~\ref{tab:real-robot-results} at the end of this supplement.

\section{ParaVLA Details}

\subsection{Visual and Language Streams}

We use two observations, the external agent view and the wrist view. Both are encoded by a shared DINOv2-Large~\cite{dinov2} backbone at 224 resolution. Camera-specific patch tokens remain explicit rather than being pooled into one global vector. The released reference trains the final four DINO blocks and the patch embedding, with reduced learning-rate multipliers, while the earlier DINO blocks remain frozen.

The real task instruction is encoded once per task by frozen T5-large. Its token-level hidden states, up to 64 tokens, are cached and projected to the policy width. Padding is masked and no sentence-level mean pooling is used. Thus ParaVLA preserves word-level task structure without updating the pretrained language encoder or incurring a T5 forward pass at every control step.

\subsection{Grounded Fusion and Action Expert}

The model represents its visual, language, and action tokens with 1024-dimensional feature vectors. A pre-fusion module is retained for architectural compatibility, but its vision--language fusion gate remains disabled throughout training and evaluation. Consequently, this module does not combine the visual streams with the projected T5 tokens. Active visual--language fusion occurs in the action head: the flow-matching action expert contains eight 16-head blocks in which action hidden states receive visual and text information through distinct attention computations rather than through an undifferentiated concatenated prefix. State conditioning is applied through the expert's state-conditioning path.

The expert predicts a 16-step action chunk and executes eight actions before replanning. Inference uses 20 flow-integration steps. Temporal ensembling is enabled with coefficient 0.9. The training objective is the standard masked flow-matching velocity loss; no paraphrase pair, paraphrase consistency loss, or LIBERO-Para instruction is used in training.

\par\medskip
\noindent\begin{minipage}{\columnwidth}
\centering
\footnotesize
\begin{tabularx}{\columnwidth}{@{}rXX@{}}
\toprule
Task & Canonical instruction & OOD paraphrase \\
\midrule
1 & Use your right arm to pick up the larger red object. & Use your right arm to grasp the larger red object. \\
2 & Use your right arm to pick up the leftmost cube. & Use your right arm to grasp the cube that is leftmost. \\
3 & Use your right arm to pick up the rightmost cube. & Use your right arm to pick up the cube that is rightmost. \\
4 & Use your right arm to pick up the blue cylinder. & Use your right arm to grasp the blue cylinder. \\
5 & Use your right arm to stack the leftmost cube onto the larger red object. & Use your right arm to place the leftmost cube onto the larger red object. \\
6 & Use your right arm to stack the leftmost cube onto the smaller blue object. & Use your right arm to place the leftmost cube onto the smaller blue object. \\
\bottomrule
\end{tabularx}
\captionof{table}{Canonical instructions and unseen OOD paraphrases used in the real-robot evaluation.}
\label{tab:supp-24}
\end{minipage}

\par\medskip
\noindent\begin{minipage}{\columnwidth}
\centering
\footnotesize
\begin{tabularx}{\columnwidth}{@{}r*{4}{>{\centering\arraybackslash}X}@{}}
\toprule
Task & Native C & Native OOD & GSR C & GSR OOD \\
\midrule
1 & 0\% & 0\% & 100\% & 100\% \\
2 & 0\% & 0\% & 100\% & 40\% \\
3 & 0\% & 0\% & 0\% & 0\% \\
4 & 0\% & 0\% & 100\% & 100\% \\
5 & 0\% & 0\% & 0\% & 0\% \\
6 & 0\% & 0\% & 0\% & 0\% \\
\textbf{SR} & \textbf{0\%} & \textbf{0\%} & \textbf{50\%} & \textbf{40\%} \\
\bottomrule
\end{tabularx}
\captionof{table}{Repeated real-robot evaluation with five trials per task and instruction condition. C denotes the canonical instruction and OOD denotes the unseen paraphrase.}
\label{tab:real-robot-results}
\end{minipage}

\clearpage
\bibliographystyle{plainnat}
\bibliography{main}
\end{document}